\documentclass[10pt,twocolumn,letterpaper]{article}

\usepackage{cvpr}              

\usepackage{graphicx}
\usepackage{multirow}
\usepackage{makecell}
\usepackage{float}
\usepackage{subcaption}

\usepackage[accsupp]{axessibility}  
\definecolor{cvprblue}{rgb}{0.21,0.49,0.74}
\usepackage[pagebackref,breaklinks,colorlinks,allcolors=cvprblue]{hyperref}

\def\paperID{35} 
\def\confName{CVPR}
\def\confYear{2025}

\title{Geometric Consistency Refinement for Single Image Novel View Synthesis via Test-Time Adaptation of Diffusion Models}

\author{Josef Bengtson, David Nilsson and Fredrik Kahl\\ 
Computer Vision Group, Chalmers
University of Technology \\
{\tt\small \{bjosef,david.nilsson,fredrik.kahl\}@chalmers.com}
}

\begin{document}
\maketitle
\begin{abstract}
Diffusion models for single image novel view synthesis (NVS) can generate highly realistic and plausible images, but they are limited in the geometric consistency to the given relative poses. The generated images often show significant errors with respect to the epipolar constraints that should be fulfilled, as given by the target pose. In this paper we address this issue by proposing a methodology to improve the geometric correctness of images generated by a diffusion model for single image NVS. We formulate a loss function based on image matching and epipolar constraints, and optimize the starting noise in a diffusion sampling process such that the generated image should both be a realistic image and fulfill geometric constraints derived from the given target pose. Our method does not require training data or fine-tuning of the diffusion models, and we show that we can apply it to multiple state-of-the-art models for single image NVS.
The method is evaluated on the MegaScenes dataset and we show that geometric consistency is improved compared to the baseline models while retaining the quality of the generated images.
\end{abstract}

\section{Introduction}

\begin{figure}[t]
    \centering
    \includegraphics[width=\linewidth]{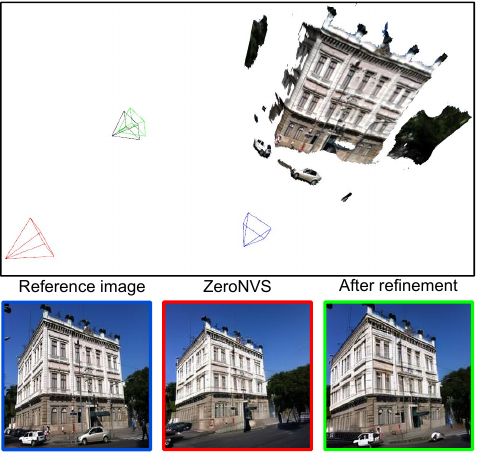}
    \caption{The single image novel view synthesis task is to, given a reference image and a relative pose, generate an image of the scene from the target pose. The estimated pose for an image generated by the diffusion based method ZeroNVS is shown in red and our refined estimate is depicted in green. The reference pose is shown in blue and the target pose in black. As can be seen, the estimated relative poses from the image generated by the diffusion model can differ significantly from the target pose. Our method refines such images to better align with the target pose.}
    \label{fig:intro_megascenes_poses}
\end{figure}

\begin{figure*}[t]
    \centering
    \begin{subfigure}[t]{0.63\linewidth}
        \centering
        \includegraphics[width=\linewidth]{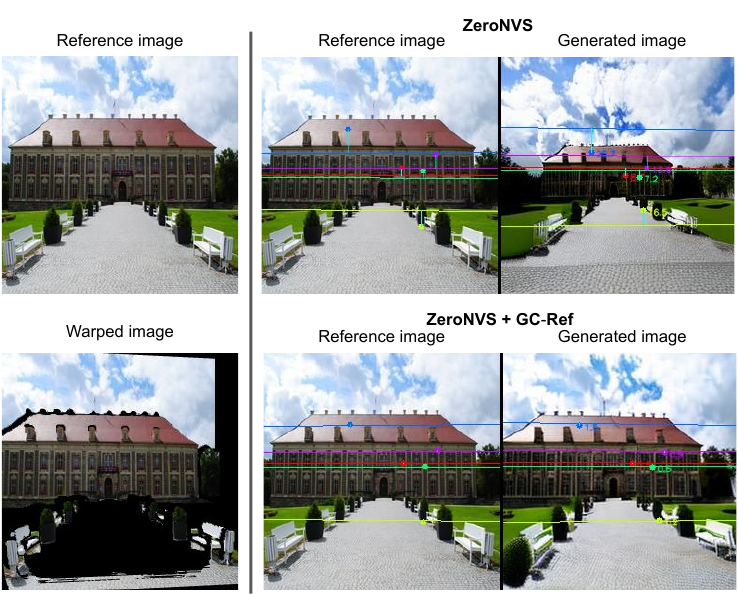}
    \end{subfigure}
    \begin{subfigure}[B]{0.36\linewidth}
        \centering
        \includegraphics[width=\linewidth]{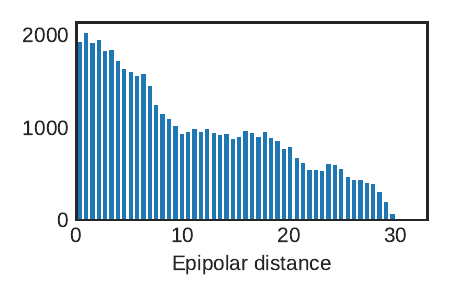}
        \vspace{-0.75cm}
        \includegraphics[width=\linewidth]{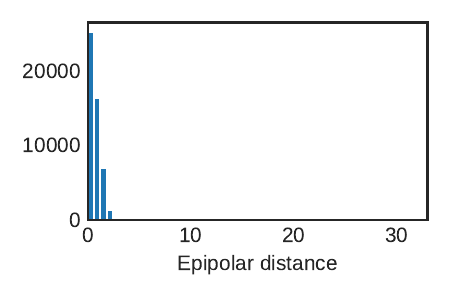}
        \label{fig:epipolar_histograms}
    \end{subfigure}
    \caption{Our method for geometric consistency refinement (GC-Ref) modifies images such that corresponding points in the reference image and the generated image lie close to their corresponding epipolar lines. We show an example of a reference image with a warping to the target pose, obtained via monocular depth estimation, that the generated image should align with. If we consider matching points between the reference images and the generated images we see that after our refinement the points lie closer to their epipolar lines. This is also shown in the histograms where we show the distributions of the distances between matching points and their corresponding epipolar lines before and after our refinement.}
    \label{fig:intro_epipolar_lines}
\end{figure*}

Given a single or multiple views of a scene, novel view synthesis (NVS) is the task of generating a new image of the scene from another viewpoint. If we are given multiple views that densely cover the full scene, this is largely a reconstruction and interpolation problem, with NeRF \cite{mildenhall2021nerf} and Gaussian Splatting \cite{kerbl20233d} being two successful approaches. However, given just a single image of a scene the problem is about generating a plausible image from the new viewpoint and there can be multiple different images that are realistic images of the scene from the new viewpoint since large parts of the generated image might not be visible in the given reference image. For this reason the single image version of the novel view synthesis problem is typically approached with generative methods, and especially diffusion models have achieved promising results on the task. 
Current state-of-the-art methods for single image novel view synthesis typically fine-tune pre-trained diffusion models on large scale datasets while changing the architecture to add the reference image and relative pose as conditioning. The pioneering work Zero-1-to-3 \cite{liu2023zero} refined a pre-trained text-conditioned image diffusion model \cite{rombach2022high} for single image novel view synthesis. This was done by changing the conditioning from text to pose and refining the model on a large scale dataset with 3D objects \cite{deitke2023objaverse}. However, it was limited to single objects and inward facing poses on a sphere. Later, this was extended to general scenes and non-restricted relative poses by ZeroNVS \cite{sargent2024zeronvs}. This was further scaled up by the introduction of the MegaScenes dataset \cite{tung2025megascenes} where a diffusion model was trained with the larger dataset, and the model architecture was also improved by adding pose conditioning via warping of the reference view to the target pose via monocular depth estimation \cite{yang2024depth}.

While these approaches are mostly capable of generating realistic and plausible images, the images do not always adhere to the target poses, as seen in Fig. \ref{fig:intro_megascenes_poses} where we show a generated image along with the target pose used as conditioning and with the relative pose estimated from the generated images. We see that the estimated poses can deviate from the target pose that the diffusion model is conditioned on. We hypothesize that this is due to the indirect encoding of the target pose to the diffusion models. In Zero-1-to-3 \cite{liu2023zero} and ZeroNVS \cite{sargent2024zeronvs} the relative pose, either given via spherical coordinates or the pose matrix, is flattened and concatenated with the CLIP embedding \cite{CLIP} of the reference image, which is then used in the cross-attention layers of the diffusion model. In MegaScenes \cite{tung2025megascenes} the pose conditioning is extended to also include a warping of the reference view to the target pose via estimated monocular depth. While this increases the geometric preciseness, it can still be observed from generated images that they are not precisely seen from the correct relative pose, and can still deviate from the warping. It is also worth noting that the warping is resized by $1/8$th to the dimensions of the latent space of the diffusion model, which removes fine details and obscures their precise location in the image.

The main idea for our proposed method is to modify the generated image such that corresponding points between the reference image and the generated image better fulfill epipolar constraints. In existing single image NVS methods there is typically a large displacement between the matching points and the corresponding epipolar lines, and our proposed method can modify the generated images such that the epipolar distances are reduced, as can be seen in Fig. \ref{fig:intro_epipolar_lines}.

Our proposed method uses pre-trained diffusion models, but changes the sampling procedure to simultaneously minimize a geometric consistency loss based on epipolar geometry. See Fig. \ref{fig:modelfig}. We generate a first candidate image by sampling random noise and denoising it with a diffusion model. We then compute matching points between the reference image and the generated image and define a loss function based on the distances between matching points and their corresponding epipolar lines. The initial noise of the diffusion process is optimized to decrease this loss, and as a result the generated image will better adhere to the epipolar geometry. Our method is used only at the generation stage, and the diffusion model does not require any further training.
In summary, our contributions are:
\begin{itemize}
    \item We introduce a training-free methodology that improves the geometric consistency of images generated by diffusion models for single image novel view synthesis.
    \item We test our method on the MegaScenes dataset, and show that we can improve the geometric correctness of several state-of-the-art models for single image NVS.
\end{itemize}

\section{Related Work}
In this section we review work on single image novel view synthesis and various approaches for 3D consistent novel view synthesis. We also present different approaches for diffusion model guidance. 

\subsection{View Conditioned Diffusion Models}

For NVS using diffusion models it is common to start with a foundation model \cite{rombach2022high} for image generation and fine-tune it for the NVS task by adding suitable conditioning in the form of some pose encoding or warping via monocular depth. 
Zero-1-to-3 \cite{liu2023zero} first trained a NVS diffusion model on a large-scale dataset \cite{deitke2023objaverse} of synthetic 3d objects. 
3DiM \cite{watson2023novel} similarly added pose-conditioning to a diffusion model and proposed sampling strategies to generate multiple consistent views.
ZeroNVS \cite{sargent2024zeronvs} refined Zero-1-to-3 using a mixture of datasets containing images from real scenes, and it was not restricted to poses on a sphere around a single object, instead being able to handle arbitrary camera transformations.
MegaScenes \cite{tung2025megascenes} trained a diffusion model similar to ZeroNVS but with additional training on the larger MegaScenes dataset and with extra conditioning using warpings based on estimated monocular depth of the reference view. For these models the main improvements come from additional fine-tuning on large scale datasets, while we instead investigate improvements that do not require additional training, and instead explicitly optimize for geometric consistency at test-time.

\subsection{Improving 3D Consistency}
There exists earlier work on improving the geometric consistency for view-conditioned diffusion models by adding additional layers that should encourage consistency \cite{ConsistNet,Consistent123,Consistent-1-to-3, free3D}. These allow for improved 3D consistency by including multi-view and epipolar guidance in the diffusion sampling process, but require additional significant fine-tuning. These methods are only trained on datasets containing single objects, and would require additional fine-tuning to be able to use on general scenes. In contrast our method can be used directly to improve performance on general scenes without requiring additional training. Another approach is \cite{viewfusion} which is an auto-regressive method that utilizes previously generated views as context when generating a sequence of new views. This method does not require additional fine-tuning and improves multiview consistency for a generated sequence, however it does not address the issue of the geometric consistency of a single generated view. Our method instead focuses on ensuring that generated images are geometrically aligned with the provided target pose.

\begin{figure*}[t]
    \centering
    \includegraphics[width=1\linewidth]{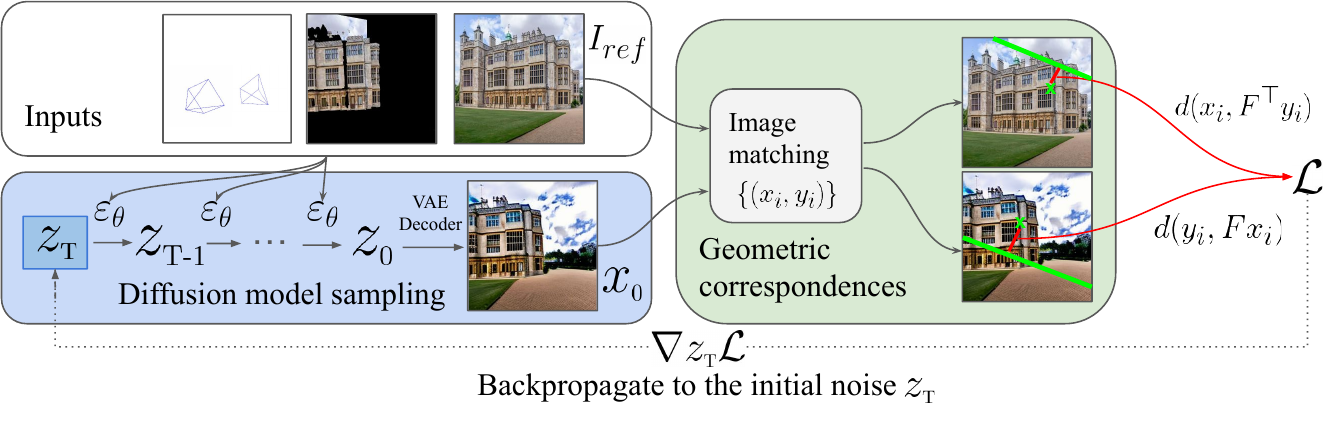}
    \caption{Our method for geometric consistency refinement aims to iteratively refine an image $x_0$ generated by diffusion model $\epsilon_\theta$ for single image view synthesis to better fulfill geometric constraints. Our method is based on the fact that if the images are geometrically consistent given the target pose then all matching points between the reference image $I_{ref}$ and the generated image $x_0$ should lie on the corresponding epipolar lines. We explicitly optimize this criteria by computing matching points between the reference image and generated image via a differentiable matcher and then use the epipolar distances as a loss function $\mathcal{L}$ to optimize the starting noise $z_T$ of the diffusion process using the gradient $\nabla_{z_T} \mathcal{L} $.}
    \label{fig:modelfig}
\end{figure*}

\subsection{3D Generation}

An orthogonal approach to directly generating novel views via a diffusion model is to use a 3D model, such as a NeRF \cite{mildenhall2021nerf}, as an intermediate representation. This is commonly done with Score Distillation Sampling (SDS) as introduced in DreamFusion \cite{poole2023dreamfusion} where SDS is used as a loss function to train a NeRF model such that renderings are similar to samples from the diffusion model. Renderings from 3D models trained with SDS are by design geometrically correct, however there are often artifacts such that the backgrounds have a uniform color and no details, and that the images are oversaturated. Another issue is that creating the 3D model requires significant compute, typically in the order of hours for a single scene. DreamFusion considered only text-to-3D generation, but this has since been extended to single image novel view synthesis \cite{melas2023realfusion,wu2024reconfusion,liu2024one,ye2024consistent}. A different approach is to train a large reconstruction model \cite{LRM} that directly predicts a 3D model of an object from a single input image, however the models can not handle general scenes. Our focus is on improving geometric consistency of view conditioned diffusion models capable of handling general scenes without requiring additional training. 

\subsection{Guidance for Diffusion Models}
To adapt a diffusion model for a specific task there are in general two approaches, either to fine-tune the diffusion model or to modify the sampling process. Fine-tuning is commonly done using e.g. LoRa \cite{hu2021lora} or ControlNet \cite{zhang2023adding} which requires ample training data and compute for training. The other approach is to guide the diffusion sampling process to steer it in a desired direction. Instead of training data this typically requires some loss function or criteria that should be minimized by the generated images. In Classifier guidance \cite{dhariwal2021diffusion} a correction is added to the noise prediction by a diffusion model to steer the sample in the direction of the negative gradient of a loss function. The loss function can e.g. be an image classifier or segmentation model. In a similar manner there exists training-free guidance methods \cite{universalGuidance,TFG,FreeDoM} that add corrections using a one-step prediction of the clean image from the current noisy sample to compute the guidance. Another approach is to optimize the initial noise for the diffusion process \cite{Samuel2023NAO,Samuel2023SeedSelect}, where the initial noise is optimized such that the generated images are semantically similar to a set of reference images.

\section{Methodology}
In this section we present our geometric consistency loss followed by a description of how we use this loss in the diffusion sampling process to generate more geometrically consistent images. 

\subsection{Geometric Consistency Loss}
Our goal is to generate an image that precisely matches the relative pose given as input, and  we will here present the loss function used to achieve this. Our main criteria for measuring geometric consistency are the epipolar distances, using the fact that matching points should lie on their corresponding epipolar lines. We will use a differentiable matching method to obtain dense matches between the reference image and the currently generated image and optimize the image to get lower distances to epipolar lines. An overview is shown in Fig. \ref{fig:modelfig}.

We obtain multiple matches $M = \{(x_i, y_i)\}_{i=1}^N$ where $x_i$ is in the reference image and $y_i$ in the generated image using the RoMa matcher \cite{edstedt2024roma}, from which we obtain a dense set of matches, along with confidence scores which are used to remove uncertain matches. We define the geometric consistency loss as the average distance of matches $(x_i, y_i)$ to their corresponding epipolar lines $Fx_i$ and $F^T y_i$ where $F$ is the fundamental matrix obtained from the target pose. The full loss $\mathcal{L}$ is defined as
\begin{align}
    \mathcal{L} = \frac{1}{N} \sum_{i=1}^N \left( \rho \left( d(y_i, Fx_i) + d(x_i, F^T y_i) \right) + \right. \\
    \left. \lambda_{rgb} \| I_{ref}(x_i)-I_{gen}(y_i)\|_1 \right) \notag
\end{align}
where $I_{ref}$ is the reference image, $I_{gen}$ is the current generated image and $\rho$ is a robust Huber loss with a threshold of $2$ pixels. The first term enforces matching points to lie close to their corresponding epipolar lines and the second term is a photo-consistency RGB loss enforcing matching points to have similar colors, and this loss term is weighted by a constant $\lambda_{rgb}$.

\begin{table*}[]
    \centering
    \begin{tabular}{l c c c c c c}\hline
         & \(\downarrow R_{dist}\) & \(\downarrow T_{dist}\) & $\uparrow$ \makecell{Masked \\ PSNR} &$\uparrow$ \makecell{Masked \\ SSIM} & $\downarrow$\makecell{Masked \\ LPIPS} & $\downarrow$\makecell{Masked \\ FID}  \\ \hline
        MegaScenes & 3.70 & 11.30 & 16.97 & 0.595 & 0.213 & 72.97 \\
        MegaScenes + GC-Ref & \textbf{2.88} & \textbf{7.94} & \textbf{18.13} & \textbf{0.631} & \textbf{0.199} & \textbf{68.48}\\
        \hline
        ZeroNVS-MS & 7.04 & 31.95 & 14.15 & 0.512 & 0.303 & 92.27\\
        ZeroNVS-MS + GC-Ref & 5.73 & 20.82 & 14.82 & 0.536 & 0.276 & 85.53\\

    \end{tabular}
    \caption{Main results of our geometric consistency refinement (GC-Ref). We see that for both MegaScenes and ZeroNVS the images better align with the target poses after our refinement, as shown by the rotation and translation errors $R_{dist}$ and $T_{dist}$ decreasing. For the reconstruction metrics we also see a similar trend, namely that with our refinement the images are closer to the warping of the reference images, indicating improvements in both geometric preciseness and image quality.}
    \label{tab:main_results}
\end{table*}

We filter the matches used in the computation of the loss $\mathcal{L}$ based on their confidence scores, and use fix positions of the matches in the reference image throughout the optimization. Given the starting noise $z_T$ we obtain the initial generated image $I_{gen}^0$ from which we compute matches $M = \{(x_i, y_i)\}_{i=1}^N$ where $x_i$ is in the reference image $I_{ref}$ and $y_i$ is in $I_{gen}^0$. For the loss $\mathcal{L}$ we use all matches with confidence scores above a threshold, which is a hyperparameter we select. During the optimization process at iteration $t$ we recompute the matches between $I_{ref}$ and the current generated image $I_{gen}^t$, and we use the matches $\{(x_i, \tilde{y}_i)\}_{i=1}^N$ where $x_i$ is kept from the initial match filtration and $\tilde{y}_i$ is its current match in $I_{gen}^t$. We compare several alternative ways of filtering the matches in Sec. \ref{sec:ablations}.

 \subsection{Image Generation}

We use diffusion models pre-trained for single image novel view synthesis, and our method refines the generated images to better align with the geometry specified by the relative pose, as measured by the loss described in the previous section.
To refine the generated images we use an approach inspired by SeedSelect \cite{Samuel2023NAO,Samuel2023SeedSelect}, as shown in Fig. \ref{fig:modelfig}. We use deterministic diffusion model sampling with DDIM \cite{songdenoising}, and optimize the starting noise. If the starting noise is $z_T$, we run the diffusion model for $T=50$ steps to get $z_{T-1}, \ldots, z_0$, in each step evaluating the diffusion model $\epsilon_\theta(z_t; t,I_{ref},c)$, where $c$ contains conditioning regarding target pose. The latent $z_0$ is then decoded to the image space $x_0$, and the geometric matching loss $\mathcal{L}$ is computed using the clean image $x_0$ and the given reference image $I_{ref}$. The starting noise $z_T$ is optimized using the gradient $\nabla_{z_T} \mathcal{L} $. We use $35$ iterations to refine $z_T$, with the Adam optimizer \cite{Adam} and the learning rate $0.025$.

\section{Experiments}
In this section we present experiments using our  training-free methodology for improved geometric consistency for single image NVS. 

\subsection{Setup}
\begin{figure}[t]
    \centering
    \includegraphics[width=1\linewidth]{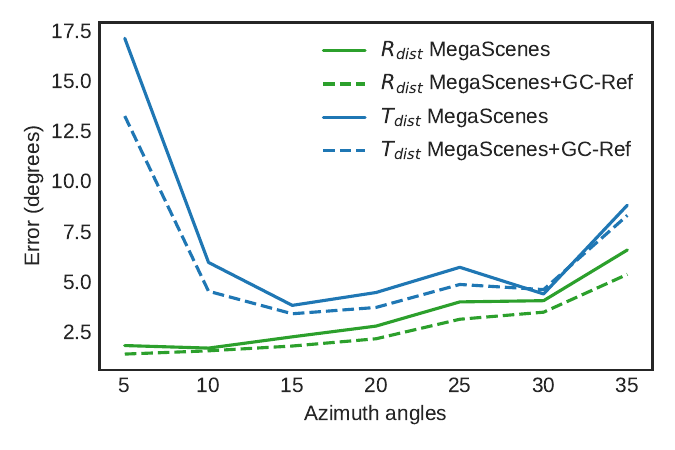}
    \caption{Given the reference image and generated image we estimate the relative poses and compare it with the target poses. We show how the pose errors vary with the magnitude of the rotation of the target pose. The rotation errors increase the more the target pose is rotated from the reference pose, while the translation errors are largest for small camera motions. For both measures we see that with our geometric consistency refinement (GC-Ref) the errors are reduced compared to the baseline methods.}
    \label{fig:plot_poseerr_vs_azim}
\end{figure}

\paragraph{Dataset.} We evaluate our method on the MegaScenes dataset \cite{tung2025megascenes}. The target pose is normalized in the same convention as for Megascenes. The estimated monocular depth is aligned with the COLMAP reconstructions \cite{schonberger2016structure} and normalized such that the 20th percentile of the depth of the reference has unit length. We follow MegaScenes \cite{tung2025megascenes} and compute monocular depth of the reference views with DepthAnythingV2 \cite{yang2024depth}, and create a mesh by unprojecting the depth and creating mesh faces from neighbouring pixels. Surfaces in the mesh are removed if the angle between the surface normal and the direction to the center of the reference camera is above a certain threshold, which we set to $70^{\circ}$. The images in the MegaScenes dataset come from Internet photo collections and cover a wide variation of scenes (plazas, buildings, interiors and natural landmarks) captured under varying conditions (differing lighting, weather and camera intrinsics). We choose a subset of images capturing some of these variations, while filtering out images where failed monocular depth estimation leads to incorrect warpings. This resulted in 46 images from the validation set, that are used for hyperparameter tuning and the ablation study, and 113 images from the test set, which are used for the final evaluation. 
We evaluate our method using 14 target poses generated using spherical coordinates, changing relative angles of the target pose with azimuth and elevation angles respectively in the range  $\left [ 5^{\circ}, 10^{\circ},\ldots , 35^{\circ} \right ]$. Leading to a total of 644 images for validation and 1582 images for testing.

\begin{table*}[]
\centering\begin{tabular}{lllcccccc}
                                                                           &                                                                             &                           & \(\downarrow R_{dist}\)        &  \(\downarrow T_{dist}\) & $\uparrow$ \makecell{Masked \\ PSNR} &$\uparrow$ \makecell{Masked \\ SSIM} & $\downarrow$\makecell{Masked \\ LPIPS} & $\downarrow$\makecell{Masked \\ FID} \\ \Xhline{4\arrayrulewidth}
                                                                           & \begin{tabular}[c]{@{}l@{}}Without Optimization\end{tabular}              & \multicolumn{1}{l|}{}     & 3.38          & 8.21           & 17.00                                                 & 0.629                                                  & 0.205                                                  & 73.52                                                \\ \Xhline{4\arrayrulewidth}
\multirow{7}{*}{\begin{tabular}[c]{@{}l@{}}Match\\ Filtering\end{tabular}} & No Filtering                                                                & \multicolumn{1}{l|}{}     & 3.50          & 8.64           & 17.33                                                 & 0.641                                                  & 0.200                                                  & 72.34                                                \\ \cline{2-9} 
                                                                           & \multirow{3}{*}{\begin{tabular}[c]{@{}l@{}}Fix\\ Matches\end{tabular}}      & \multicolumn{1}{l|}{0.05} & 2.70          & 5.62           & 18.28                                                 & 0.660                                                  & 0.192                                                  & \textbf{69.75}                                       \\
                                                                           &                                                                             & \multicolumn{1}{l|}{0.15} & \textbf{2.64} & \textbf{4.95}  & \textbf{18.43}                                        & \textbf{0.664}                                         & \textbf{0.190}                                         & 69.83                                                \\
                                                                           &                                                                             & \multicolumn{1}{l|}{0.25} & 2.81          & 7.10           & 18.39                                                 & 0.662                                                  & \textbf{0.190}                                         & 70.30                                                \\ \cline{2-9} 
                                                                           & \multirow{3}{*}{\begin{tabular}[c]{@{}l@{}}Adaptive\\ Matches\end{tabular}} & \multicolumn{1}{l|}{0.05} & 2.68          & 7.42           & 18.10                                                 & 0.660                                                  & 0.196                                                  & 71.79                                                \\
                                                                           &                                                                             & \multicolumn{1}{l|}{0.15} & 2.87          & 6.43           & 18.28                                                 & \textbf{0.664}                                         & 0.191                                                  & 70.75                                                \\
                                                                           &                                                                             & \multicolumn{1}{l|}{0.25} & 3.20          & 7.88           & 18.29                                                 & \textbf{0.664}                                         & 0.191                                                  & 70.04                                                \\ \Xhline{4\arrayrulewidth}
\multirow{3}{*}{\begin{tabular}[c]{@{}l@{}}RGB\\ Loss\end{tabular}}        & \multirow{3}{*}{\(\lambda_{\text{rgb}}\)}                             & \multicolumn{1}{l|}{0}    & 2.79          & 6.71           & 15.88                                                 & 0.613                                                  & 0.214                                                  & 78.28                                                \\
                                                                           &                                                                             & \multicolumn{1}{l|}{2.5}  & \textbf{2.64} & \textbf{4.95}  & 18.43                                                 & 0.664                                                  & 0.190                                                  & 69.83                                                \\
                                                                           &                                                                             & \multicolumn{1}{l|}{10}   & 2.66          & 6.85           & \textbf{18.62}                                        & \textbf{0.667}                                         & \textbf{0.187}                                         & \textbf{68.79}                                      
\end{tabular}

\caption{Ablation studies for how to filter matches and how to select the confidence threshold for the points we use in the loss function of our method, and also the results for different weights for the RGB loss. We found that the best performance was obtained by fixing the matches from the first generated image and to only use matches where the confidence score was at least $0.15$. Using either all matching points regardless of their confidence or using a higher confidence threshold resulted in worse performance. Additionally, we found that including the RGB loss gave significant improvements in image quality, while a too large value for $\lambda_{\text{rgb}}$ negatively impacted the pose accuracy, leading to a choice of $\lambda_{\text{rgb}}=2.5$.}
\label{tab:Ablation}
\end{table*}

\begin{figure*}[t]
    \centering
    \includegraphics[width=0.46\linewidth]{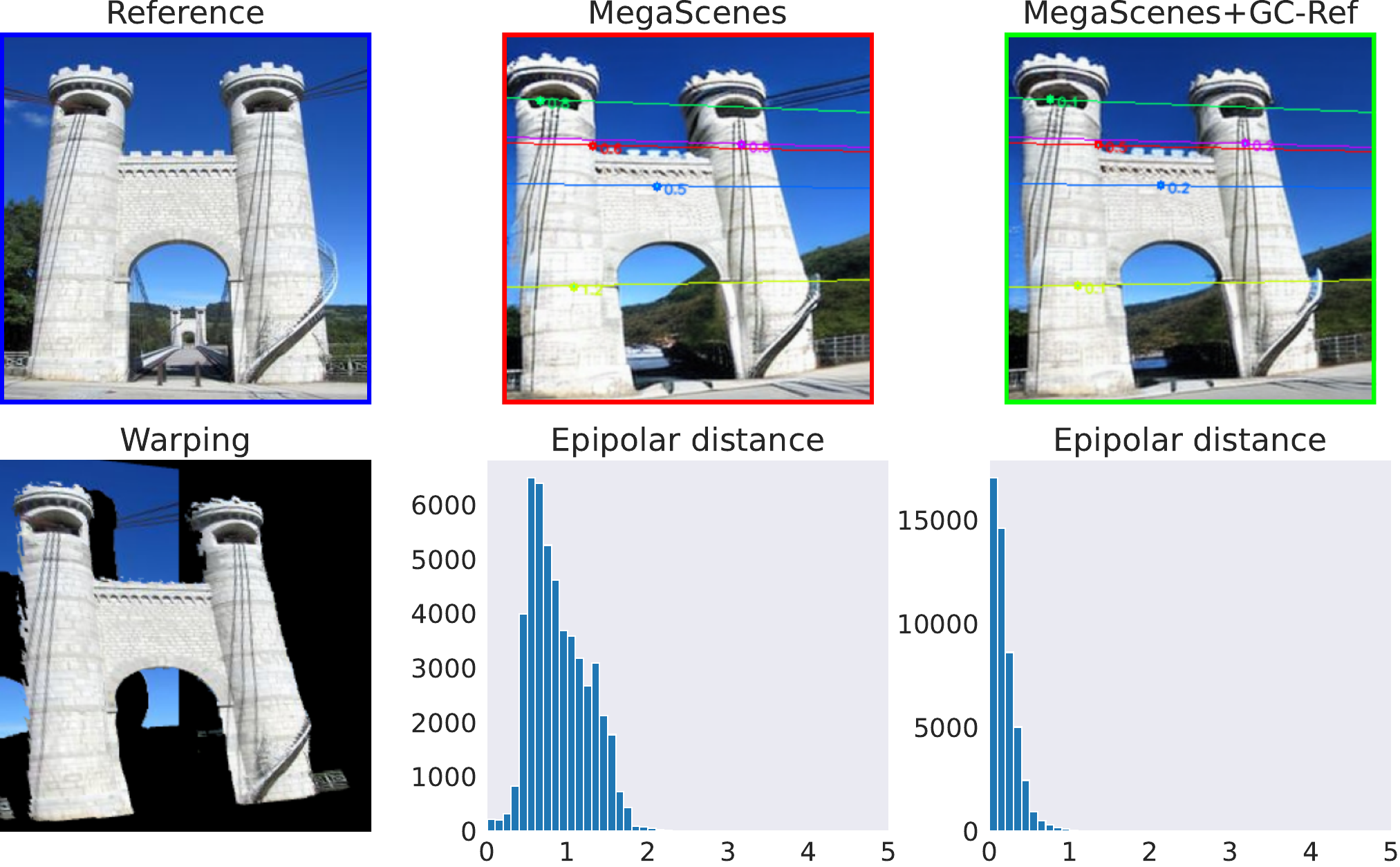}
    \includegraphics[width=0.53\linewidth]{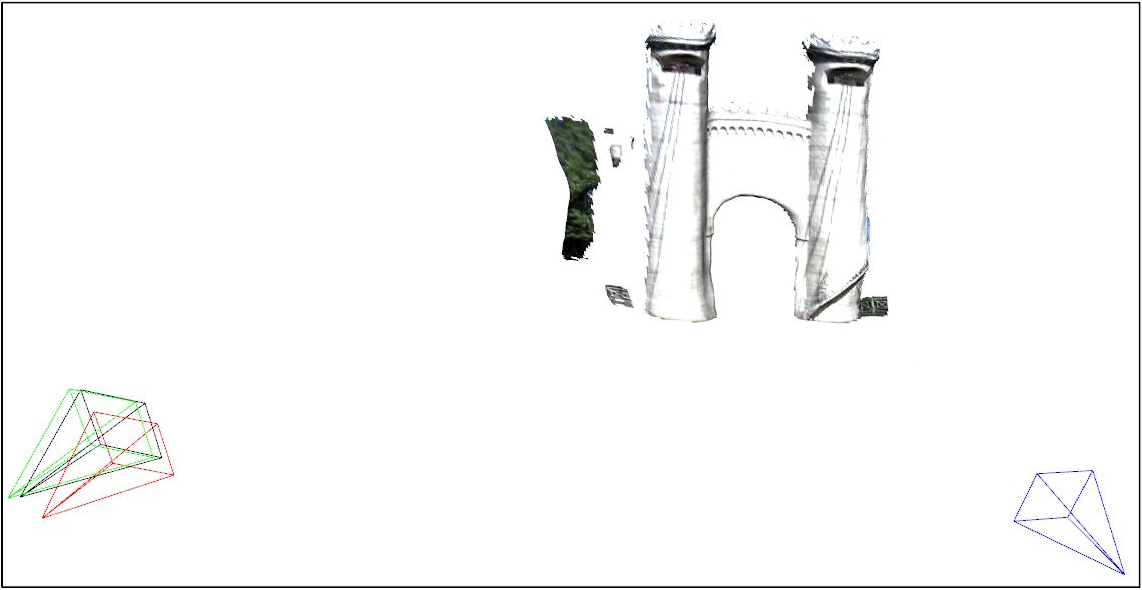}
    \caption{Our refinement leads to reduced epipolar distances and to improved pose accuracy. We see in the histogram that the distances to the epipolar lines decrease after our refinement and also in the 3D plot that the estimated camera pose (green) is closer to the target pose (black) than the original generated image (red). In this specific case the translation error decreased from $2.8^{\circ}$ to $0.8^{\circ}$ and the rotation error from $1.8^{\circ}$ to $0.8^{\circ}$.}
    \label{fig:epidist_histogram_3d_plot}
\end{figure*}

\paragraph{Metrics.} 
Following MegaScenes\cite{tung2025megascenes} we report masked reconstruction metrics and generative metrics evaluated only on the pixels that are covered by the warped reference view. For reconstruction metrics we use PSNR and SSIM \cite{SSIM} that measure similarity on pixel and patch basis and LPIPS \cite{LPIPS} that measures perceptual similarity. We also use FID \cite{FID} to evaluate the quality of generated images by comparing their feature distribution to those of real images. We do not evaluate using ground truth pairs, but with freely chosen poses, so we do not compute any metrics using ground truth target images. Additionally, to measure the geometric correctness we evaluate the pose accuracy of the generated images. Given the reference image and the generated image, we compute DeDoDe matches \cite{dedode} followed by using PoseLib \cite{PoseLib} to estimate the relative pose. Note that we used the RoMa \cite{edstedt2024roma} matcher during our optimization, so for fair comparison a different matcher is used to estimate poses for evaluation. We then measure the average rotation error $R_{dist}$ and translation error $T_{dist}$ of the estimated poses $(R_{gen}^i, T_{gen}^i)$ relative to the target poses $(R_{gt}^i, T_{gt}^i)$ by the relative angles obtained by
\begin{equation}
R_{dist} =\frac{1}{n} \sum_{i=1}^n \text{arccos}\left ( \frac{\text{tr}(R_{gen}^i R_{gt}^{i\top})-1}{2} \right)
\end{equation}
\begin{equation}
    T_{dist} = \frac{1}{n} \sum_{i=1}^n \text{arccos}\left ( \frac{T_{gen}^i \cdot T_{gt}^i}{ \left\| T_{gen}^i \right\| \left\| T_{gt}^i \right\|} \right)
\end{equation}
These two metrics capture how well the pose of the generated image match the desired target pose.

\paragraph{Implementation.} We select matches using the RoMa matcher \cite{edstedt2024roma} for the optimization. We test our methods using the ZeroNVS \cite{sargent2024zeronvs} and MegaScenes \cite{tung2025megascenes} models for single image NVS. We use the checkpoint of ZeroNVS that is fine-tuned on MegaScenes, as in \cite{tung2025megascenes}. We use an A40 GPU and the optimization takes 6 minutes per image. 

\begin{figure*}
    \centering
    \includegraphics[width=1\linewidth]{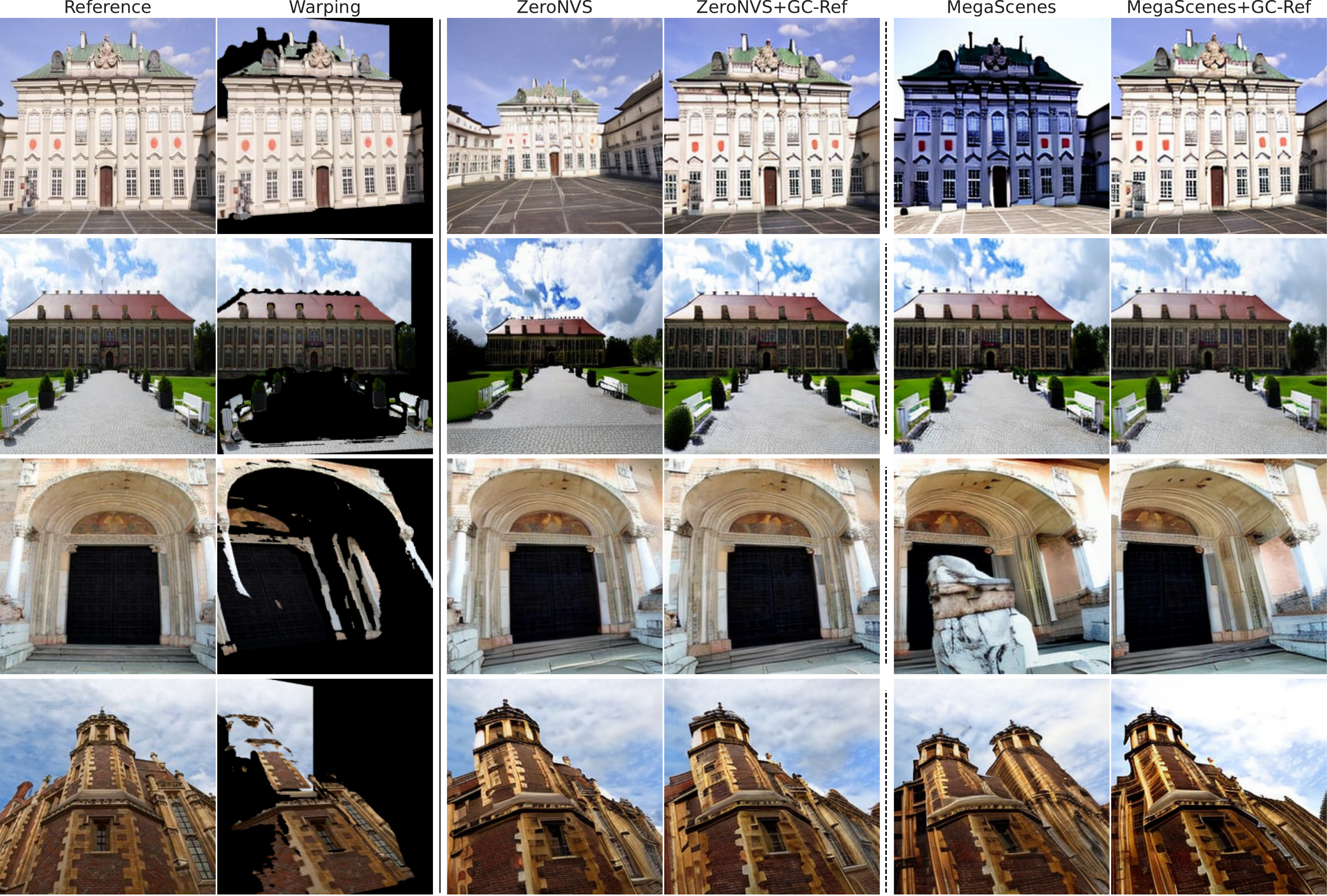}
    \caption{We show several qualitative examples of images before and after our geometric consistency refinement (GC-Ref). For images generated with ZeroNVS there is not explicit conditioning on the warping and we often see misalignment and incorrect scale, something that can be fixed by our method. The MegaScenes model conditions explicitly on the warping and it generally aligns well with the warping. However, sometimes there are errors such as incorrect lighting or hallucinated objects, which our method sometimes can resolve due to using dense matches to change large parts of the image.}
    \label{fig:qualitative_before_after}
\end{figure*}

\subsection{Main Results}
Our main results are shown in Table \ref{tab:main_results}. Our overall goal is to generate more geometrically consistent images, and we can see that our method decreases the error of the estimated relative poses, which indicates that the generated images are closer to the target pose than the images generated by the baseline methods. We also see that the generated images are more similar to the warpings than the baseline methods are, although this is partially due to the inclusion of the photo-consistency RGB loss term as shown in the ablations in Sec. \ref{sec:ablations}. Finally we see that the FID metric is improved which indicates that the generated images have more similar semantics and structure to the reference images than the baselines.

In Fig. \ref{fig:plot_poseerr_vs_azim} we show how the errors of the estimated rotations and translations depend on the relative angle of the target pose. For the rotation we see that our method obtains lower error than the baseline methods, and that the error increases with the rotation angle of the target pose. As for the translation we see that the error is largest for small camera motions. 
\begin{figure*}
    \centering
    \includegraphics[width=\linewidth]{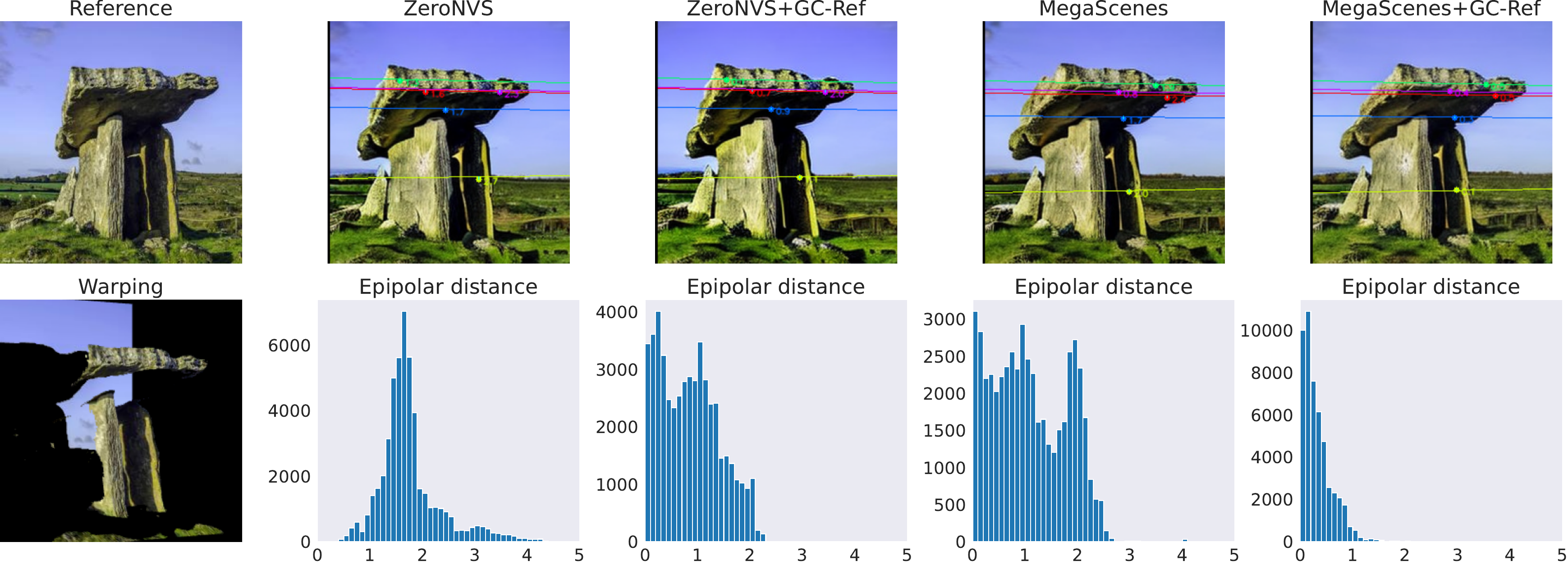}

    \vspace{0.01mm} 
    \rule{\linewidth}{0.4pt} 
    \vspace{0.01mm} 

    \includegraphics[width=\linewidth]{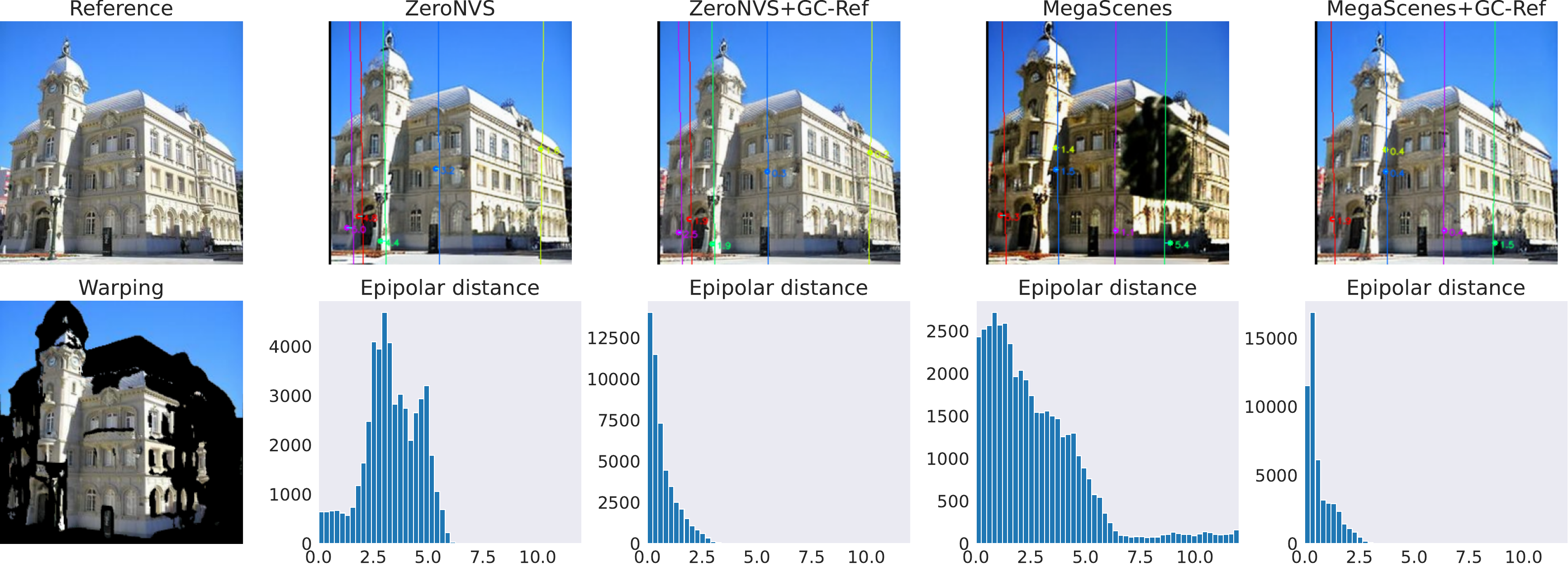}
    \caption{We show several examples of the distribution of the distances between points and their corresponding epipolar lines obtained from the target poses and dense matches. While the images often appear similar, the refined images are more geometrically consistent which we can see since the matching points are closer to the epipolar lines.}
    \label{fig:epiline-results}
\end{figure*}

We show multiple qualitative results in Fig. \ref{fig:qualitative_before_after}. For ZeroNVS, we note that the generated images often exhibit incorrect scale and do not depict the scene from the correct pose. In these cases our method can correct the generated images so that they are better aligned with the target pose and the scale is closer to that of the warping. This is similar to the example in Fig. \ref{fig:intro_epipolar_lines}. The MegaScenes model is conditioned on the warping so we do not see as drastic changes, but since we use dense matches to modify the images we sometimes observe that the images after the refinement look more similar to the reference images in terms of lighting and image content. 

In Fig. \ref{fig:epiline-results} we show the distributions of distances of matches to their corresponding epipolar lines. We see that while the images look similar the matching points are significantly closer to the corresponding epipolar lines after our refinement showing that the images better align with the target pose. Another example is shown in Fig. \ref{fig:epidist_histogram_3d_plot} where we also show the estimated camera poses in 3D. The estimated camera pose is closer to the target pose after our refinement and points lie closer to their corresponding epipolar lines, showing that our geometric consistency refinement changes the images such that they better align with the target pose. 

\subsection{Ablation Studies}\label{sec:ablations}
In Table \ref{tab:Ablation} we show ablation studies for several key design choices of our method, mainly for how to select matching points. We use RoMa \cite{edstedt2024roma} which provides a match for all pixels in the reference image along with per-pixel confidence scores. We vary both how we select the matches and the threshold at which we filter the matches. We tried using all matches or to filter based on confidence scores as given by RoMa and found that the results improved if uncertain matches were removed. Furthermore we tested both to fix the matches (in the reference image) after the first image was generated as well as to update the points in the reference image used in the optimization for every iteration.
We found that fixing the position of the matches in the reference image after the first image is generated and then keeping those matches, regardless of whether the confidence of that match increased or decreased, gave the best results.

\section{Conclusions}
We addressed the problem of geometric correctness for single image NVS. Existing methods based on diffusion models generate realistic images, but they do not always depict the scene from precisely the target poses. Our method improves the geometric consistency by refining the generated images from such models to better fulfill epipolar geometry. We evaluated our method on MegaScenes and showed that with our refinement the images better align with the target poses. Our method is training-free and does not require any fine-tuning or ground truth data, however it does require test-time optimization for each generated image.

\subsection*{Acknowledgments}
This work received full support from the Wallenberg AI, Autonomous Systems, and Software Program (WASP) funded by the Knut and Alice Wallenberg Foundation. Computational resources were provided by the National Academic Infrastructure for Supercomputing in Sweden (NAISS) at Chalmers Centre for Computational Science and Engineering (C3SE), partially funded by the Swedish Research Council under grant agreement no. 2022-06725, and by the Berzelius resource, provided by the Knut and Alice Wallenberg Foundation at the National Supercomputer
Centre.

{
    \small
    \bibliographystyle{ieeenat_fullname}
    \bibliography{main}
}


\end{document}